\documentclass[final]{cvpr}

\usepackage{times}
\usepackage{epsfig}
\usepackage{graphicx}
\usepackage{amsmath}
\usepackage{amssymb}

\newcommand{\softmax}{\mathrm{softmax}} 
\usepackage{amsmath} 
\usepackage[switch]{lineno} 
\usepackage[linesnumbered,ruled,vlined]{algorithm2e} 
\SetKwInput{KwInput}{Input}                
\SetKwInput{KwOutput}{Output}              

\usepackage[pagebackref=true,breaklinks=true,colorlinks,bookmarks=false]{hyperref}

\def\confYear{CVPR 2021}

\begin{document}
%
\title{A modular vision language navigation and manipulation framework for long horizon compositional tasks in indoor environment}

\author{Homagni Saha\\
Iowa State University\\
{\tt\small hsaha@iastate.edu}

\and
Fateme Fotouhi\\
Iowa State University\\
{\tt\small fotouhif@iastate.edu}

\and
Qisai Liu\\
Iowa State University\\
{\tt\small supersai@iastate.edu}

\and
Soumik Sarkar\\
Iowa State University\\
{\tt\small soumiks@iastate.edu}
}

\maketitle

\begin{abstract}
In this paper we propose a new framework - \textbf{MoViLan} (Modular Vision and Language) for execution of visually grounded natural language instructions for day to day indoor household tasks. While several data-driven, end-to-end learning frameworks have been proposed for targeted navigation tasks based on the vision and language modalities, performance on recent benchmark data sets revealed the gap in developing comprehensive techniques for long horizon, compositional tasks (involving manipulation and navigation) with diverse object categories, realistic instructions and visual scenarios with non-reversible state changes. We propose a modular approach to deal with the combined navigation and object interaction problem without the need for strictly aligned vision and language training data (e.g., in the form of expert demonstrated trajectories). Such an approach is a significant departure from the traditional end-to-end techniques in this space and allows for a more tractable training process with separate vision and language data sets. Specifically, we propose a novel geometry-aware mapping technique for cluttered indoor environments, and a language understanding model generalized for household instruction following. We demonstrate a significant increase in success rates for long-horizon, compositional tasks over the baseline on the recently released benchmark data set-ALFRED.

\end{abstract}

\begin{figure}
  \centering
  \includegraphics[width=0.5\textwidth]{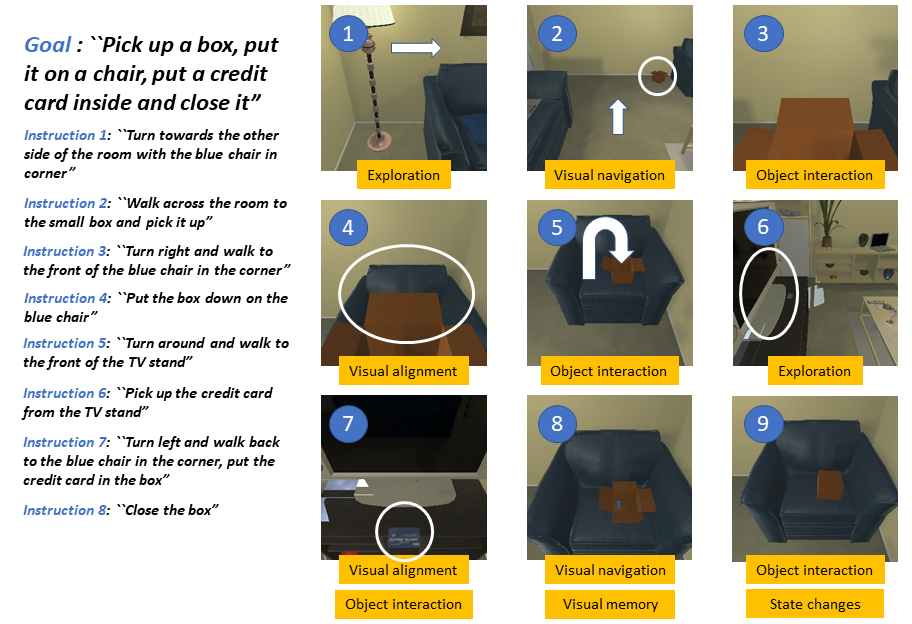}
  \caption{\textit{Example of a long horizon compositional task. An overall goal is provided to the robot in form of a sequence of instructions. Combining these with vision at current time step, it must execute actions to satisfy the goal.}}
  \label{fig:task_ex}
\end{figure}

\section{Introduction}
Vision language navigation received a lot of attention recently as its inherent cross modal nature in anchoring natural language commands to visual perception is highly relevant for practical robotic applications.
Recent progress in deep learning for visual and linguistic representations has created a push to develop techniques for ever more complex, close to real life situations involving realistic simulators, longer execution trajectories, diverse object categories and natural language descriptions. Even as newer techniques are being proposed for better visual navigation, improved language understanding, vision-language grounding, and end to end differentiable planning, there remains a lack of holistic approach to combine the elements in solving long horizon, compositional tasks with diverse object categories and irreversible state changes that require dynamic planning. Imagine having to follow a set of instructions in a visual environment as shown in figure~\ref{fig:task_ex}. While very intuitive for a human, generalization on such tasks in unseen environments have remained a hard problem. For example, a benchmark data set was proposed recently for vision-language navigation and manipulation tasks~\cite{shridhar2020alfred}. Initial study in this work achieved a very poor success rate, as low as 4 \% even when deploying state of the art learning frameworks. Several past research have tried to learn input language and expert action demonstrations into a joint action mapping using reinforcement/sequence-to-sequence learning techniques. However, judging from the nature of the language instructions used in these studies (e.g., using R2R data set), most often agents only require to identify a target location on the map and navigate to it. This situation can become much more complicated with compositional instructions common in household robotics such as-``open the microwave, put the coffee, and wait for 5 seconds''. Although the number of expected actions from the language is not immense, the complexity of this joint mapping from vision and language to action may increase drastically, even more with longer sentences. This might make existing reinforcement learning/sequence-to-sequence techniques much harder to learn. In this context, We make the following key contributions here:

\begin{figure*}
    \centering
    \includegraphics[width=0.7\linewidth]{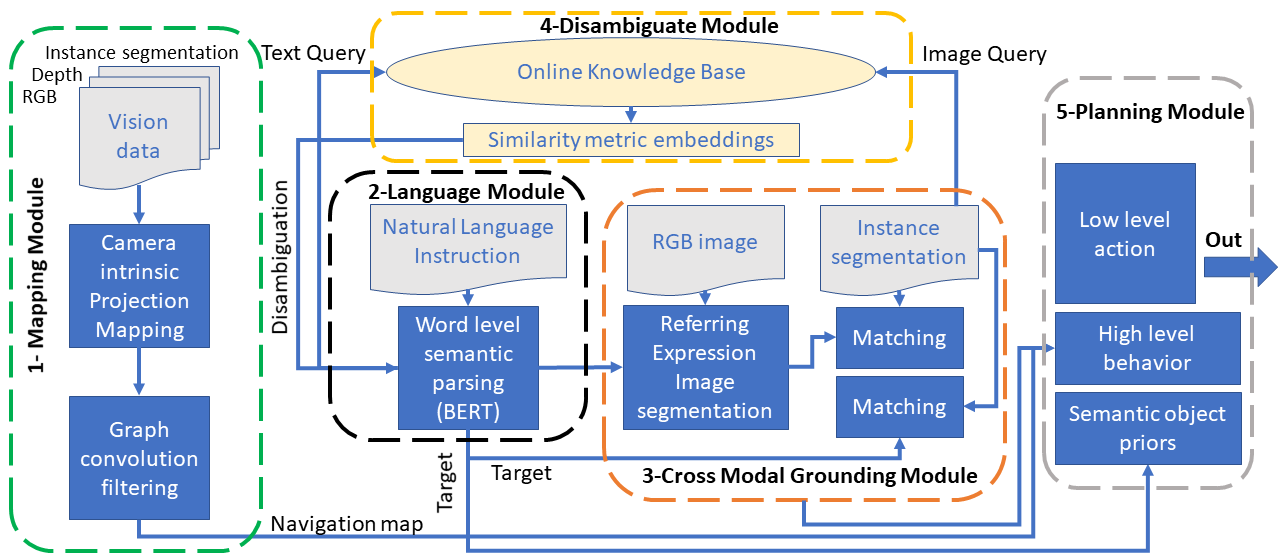}
    \caption{\textit{Overview MoViLan: All five modules are shown in dotted boxes}}
    \label{modules}
\end{figure*}

\begin{enumerate}
    \item We propose a new modular approach to deal with combined navigation and object interaction problems that involve significantly longer execution trajectories and compositionality in target objectives as well as ambiguities associated with real world language based interaction. This modular approach makes training more feasible with weakly aligned or even independently collected vision and language data.
    \item We propose a novel mapping scheme based on graph convolution networks to approximate geometry of observed objects and navigable space around an agent in cluttered indoor environments, enabling improved navigation.
    \item We propose a natural language understanding model for assigning contextual meanings to words in natural language instructions for vision language navigation. This can be obtained by leveraging pretrained architectures such as BERT using transfer learning on a small corpus that we extract for joint intent detection and slot filling.
    \item Finally, we demonstrate state-of-the-art success rate (close to 40\%) on unseen long-horizon, compositional tasks with non-reversible state changes on the new benchmark data set- ALFRED.
\end{enumerate}

\section{Related Work}
We succinctly present the key related works here. A broader literature survey is provided in the supplementary material. 

\textbf{Mapping based approaches}
In recent works for vision language navigation~\cite{anderson2019chasing}, authors use a metric map as memory architecture for navigating agents. In topological maps, the graph convention can be used to represent environments with nodes representing free space with edges between the nodes connecting the free spaces by traversable paths. 
Introduction of semantics in graph based maps can provide powerful solutions to encode relative locations of objects of interest in persistent memory~\cite{pronobis2012large,lang2014definition} and is useful when a higher level understanding of the surrounding is necessary. Deep learning approaches have also been proposed in this context, as in neural SLAM~\cite{zhang2017neural} that tries to mimic the SLAM procedures into soft attention based addressing of external memory architectures. Authors in~\cite{chaplot2020learning} also extend upon these approaches in a modular and hierarchical fashion. Several studies have also used differentiable and trainable planners for navigation~\cite{tamar2016value,khan2017end,lee2018gated}. 

Although detailed topological maps may be provided, localization in dynamic settings may not be trivial. Authors in~\cite{chen2019behavioral} explore a behavioral approach to navigation using graph convolution network over a topological map.
In this regard, our contribution is mainly constructing dense semantic topological maps from panorama images, specialized for cluttered indoor environments.

\textbf{Approaches based on language understanding}
Several techniques have been developed for interpreting user commands framed as a sequence prediction problem~\cite{mei2016listen,anderson2018vision}. Such machine level translation of natural language instructions has also been explored in the context of automatic question answering~\cite{xiong2016dynamic,seo2016bidirectional}. Following up, recently authors in~\cite{zang2018translating} uses attention mechanisms to learn alignment of language instructions with given topological map and output high level behaviors. 

In contrast, we adopt a completely unique approach that associates semantics to each word of the language instruction using a semantic slot filling model. We find that the semantics we define are easy to learn and general enough to encompass various kinds of household instructions. 
We will release the training data for this semantic slot filling model (that we use to fine-tune BERT). We believe that this will help researchers to develop advanced techniques on top of our proposed instruction simplification mechanism.

\textbf{End to end learning approaches}
Several recent deep learning approaches propose to learn a mapping directly from inputs to actions, whether structured observations are provided~\cite{mei2015talk,suhr2018situated} or the agent deals with raw visual observations~\cite{misra2017mapping,xiong2018scheduled}.
Cross modal grounding of language instructions to visual observations is often used in several works, via e.g., reinforcement learning~\cite{wang2018look,wang2019reinforced}, autoencoder architectures that impose a language instructions-based heat map on the visual observations (using U-net architectures~\cite{misra2018mapping}, attention mechanisms~\cite{zhu2020vision}, or implementation of non linear differentiable filters~\cite{anderson2019chasing}). However, as we show later in results, going end to end may not be best for generalizing to perform compositional tasks in unseen environments.

\section{Overview of MoViLan framework}

Given a sequence of natural language instructions that specify how to move in an environment and interact with surrounding objects, the goal of vision language instruction following is to have an autonomous agent complete a sequence of actions to fulfill the overall task. In figure~\ref{modules}, we provide a summary of our modular approach to solve this problem, that can be broken down into five major parts as:


    \textbf{Mapping}- Given RGB, depth and instance segmentation images (estimated from RGB), this module first constructs an explicit Birds Eye View (BEV) map of the environment around the agent by approximate projection using geometric transforms and camera intrinsic parameters. Due to inherent vulnerability of this technique with regards to noise and uncertainty, and sensitivity to camera parameters, we propose a filtering algorithm based on node classification using graph convolution. This can provide a refined egocentric map around the agent at each time step that takes into account locations of all objects around it. It also considers the navigable space discretized in unit steps corresponding to the discrete step size taken by the agent at each time step.
    
    \textbf{Language understanding}- This module takes the natural language instructions as input to the agent that is parsed in the context of the task at hand. We propose generalized semantic labels for parsing sentences which can be trained using transfer learning from BERT with as few as 1000 examples. This step helps the agent to decide targets for navigation, objects to interact with as well as relationship of the objects to each other with regards to fulfillment of the task (as understood from the language only).
    
    \textbf{Cross-modal grounding}- This module focuses on identifying regions of the input RGB image that coincide with the natural language descriptions. This is achieved by combining autoencoder based image segmentation technique with word level semantics obtained from language understanding. Using such keywords from language to identify regions in image is known as referring expression image segmentation (e.g.- ~\cite{ye2019cross}). We propose a simple restructuring of language inputs to LingUNet~\cite{misra2018mapping} and retrain it conditioned on semantic word labels provided by the language understanding module.
    
    \textbf{Disambiguation}- Often words used in the language instruction to the agent may represent object categories that are unrecognized by the image segmentation framework. However, class agnostic instance segmentation~\cite{zhang2019canet} or depth perception based techniques can still capture distinct regions in the image. We implement an algorithm that automatically downloads images from an online knowledge base (e.g., Google image search engine), and identifies the most similar segmented region in the image to the words in the instruction (using pairwise similarity embeddings created from last feature layers of the deep convolutional architectures). This module uses images extracted from the instance segmentation and words extracted from the language understanding module to generate queries. 
    
    \textbf{Planning}- The goal of this step is to combine outputs from mapping, language understanding and cross-modal grounding modules and output a sequence of low level actions. First, it takes word level semantic parsing from the language module as input. After mining for associations across a large corpus of word slot labels this module assigns ``semantic object priors" to each object identified in the language.
    This lets the agent understand for example, that small objects like ``pen" are to be picked up using ``Pickup" actions, whereas large objects like ``table" might be suited for ``Place" operation and so on (detailed algorithm provided in supplementary). Output of semantic object prior along with the navigation map from the mapping module and matched pixels from cross-modal grounding module help the agent choose a high level behavior (such as ''Pick up pen from table"). It is then converted to a sequence of low level actions using A* search algorithm with area (of the object of interest) maximization heuristic. Examples of low level actions would be- ``MoveAhead", ``RotateRight", ``PickObject", ``PutObject", etc. 
    
    \textit{\textbf{Aspect of long horizon}- A hurdle in many end to end frameworks is an attempt to combine all of the above modules we have identified into end to end planing. However our modular consideration provides us a distinct advantage as it allows to incorporate several recovery strategies out of failure modes. Consider the connection from instructions (1-6) to instruction 7 in figure \ref{fig:task_ex}. Having an independent mapping module communicating to an independent language and cross modal understanding module allows to precisely relocate the same blue chair from instruction 1 in instruction 7. Detailed layout of such strategies are provided in supplementary figure 3.}  

Next, we discuss the main technical contributions of the paper that are involved in the mapping, language understanding and cross-modal grounding modules. Due to space restrictions, details of the disambiguation and planning modules are provided in the supplementary material. 

\begin{figure*}
    \centering
    \includegraphics[width=.65\linewidth]{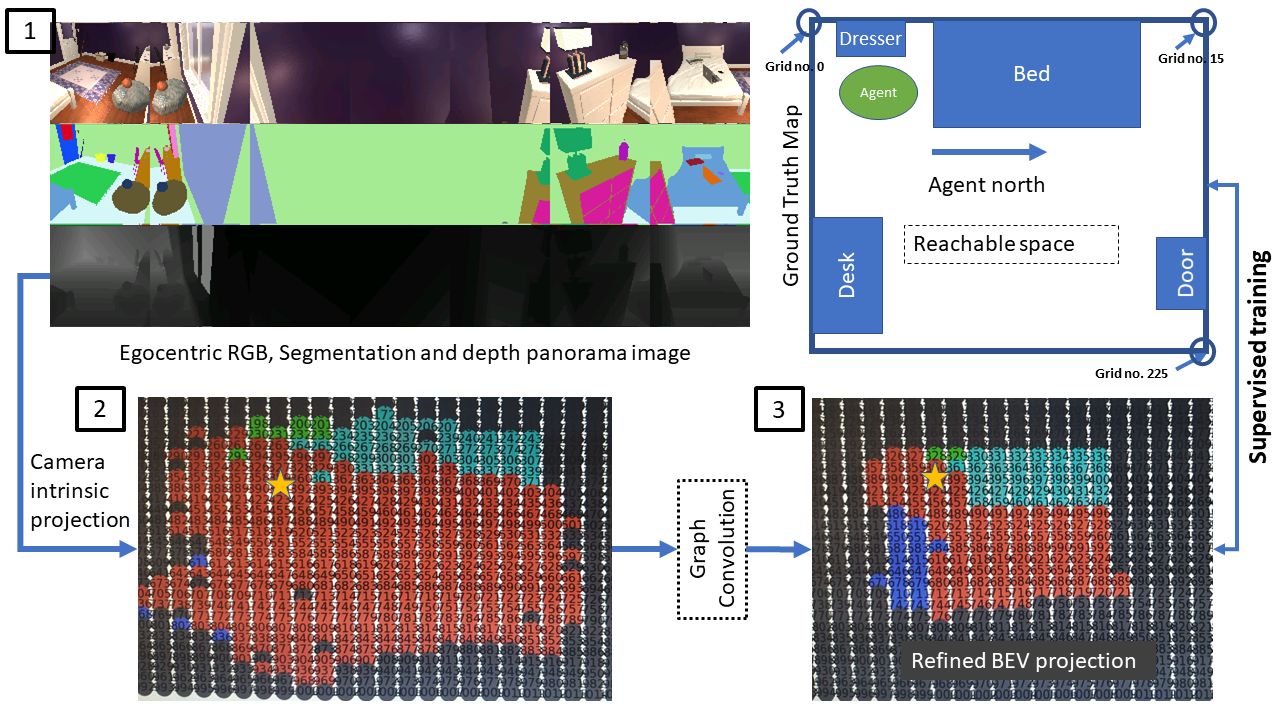}
    \caption{\textit{Schematic of mapping using graph convolution. First, the agent obtains input panorama images in RGB, segmentation and depth modalities to construct a birds eye view (BEV) projection map of objects of interest around the agent using geometric transforms with camera intrinsic parameters (e.g., Bed (cyan), Dresser (green), and Desk (blue) and the navigable space (red) are the semantic objects of interest here). Then the initial projection gets refined using a trained graph convolution network. Supervised training is used with the help of a true map (obtained by placing agent at each position in the room) for a select few rooms and agent locations. Each grid node represents 0.25m x 0.25m space. Yellow star is absolute position of the agent}}
    \label{fig:1framework}
\end{figure*}

\section{Graph convolution for mapping}
Given an RGB image $x_r \in \mathcal{R}^{h\times w\times 3}$, a corresponding segmented image $x_s \in \mathcal{R}^{h\times w\times n}$ (each pixel stores a one hot encoding denoting one out of $n$ possible object classes), and a corresponding depth normal image $x_d \in \mathcal{R}^{h\times w\times 1}$, our goal is to obtain a projection $p\in \mathcal{R}^{s\times s\times n}$. Here, $h$ is the height of the RGB image, $w$ is the width, $n$ is the number of object classes of interest, and $s$ is the size of the spatial neighborhood around the agent where we are projecting to obtain a map. Each grid of the $s\times s$ neighborhood contains a $1 \times n$ dimensional feature vector denoting one out of $n$ possible object classes occupying the grid (including navigable space). Standard image segmentation frameworks and monocular depth estimation frameworks can be used to provide $x_s$ and $x_d$ from $x_r$ with relative ease. Each pixel value of depth image can be multiplied with pixel indices, camera focal length and calibration matrix to obtain $x,y$ and $z$ distance of the pixel with respect to the agent, which is then rescaled and normalized to lie in any of the $s \times s$ grid around the agent (detail equations in supplementary). Therefore, pixel $i,j$ in $x_r,x_s$ and $x_d$ get assigned to the grid location $u(i,j)$ and $v(i,j)$ around the agent in $p\in \mathcal{R}^{s\times s\times n}$. Let $(u,v)$ be the shorthand for each grid location which now stores a $(1\times n)$ vector. Here $n$ is the number of objects of interest. Now, let each element position $l$ in the $(1\times n)$ vector store 1 if the grid location was mapped from the contributing pixel in RGB image which belonged to class $n$, and 0 otherwise. Therefore, $p_{uvl}$ can either store 1 or 0. Furthermore, as we are using panorama images, a value of $p_{uvl}$ is obtained for each discrete rotation of the agent made to complete the image (in many applications, 360 field of vision is not possible, so agent has to rotate in steps). Values $p_{uvl}$ obtained for each rotation is multiplied with rotation matrix $R(\theta)$ and averaged to obtain the final approximate projection map $p$.

\textbf{Graph convolution for projection refinement}

The graph convolution algorithm we propose is essentially a filtering algorithm for refining obtained projection elements $p_{uvl}$ in an $(s\times s)$ grid around the agent. (Fig. ~\ref{fig:1framework})

\textbf{Preliminaries on Graph Convolution Networks-}
The goal of Graph Convolution Networks (GCN) is to learn a function of signals or features on a graph $\mathcal{G} = (\mathcal{V},\mathcal{E})$. The graph $\mathcal{G}$ takes as input:

1. A description of feature $x_i$ for every node $i$, which can be summarized as an $N\times D$ feature matrix $X$, where $N$ is the number of nodes and $D$ is the number of input features.

2. A graph structure description in matrix form, supplied as an adjacency matrix $A$.

The graph $\mathcal{G}$ produces a node-level output $Z$ (an $N\times F$ feature matrix, where $F$ is the number of output features per node). Let $H^{(0)} = X$, $H^{l} = Z$, and $L$ be the number of layers of convolution, then the operation of neural network in the graph structure can be written as:
\begin{equation}
    H^{(l+1)} = f^l(H^{(l)}, A), l \in \{1,\dots,L\}
\end{equation}
Implementation of different frameworks for graph convolution chiefly differs on the choice and parameterization of the function that needs to be learned $f = \{f^1,\dots,f^l\}$. 

\textbf{Problem specific formulation-}
The graph structure we use is derived from the action space of the navigating agent in the form of a grid lattice (eg- ~\cite{zhang2019trajectory}). Entire $(s\times s)$ space around the agent is converted to a grid topology with $N = s^2$ nodes with a connection between two nodes if the Manhattan distance between them is equal to 1. Two nodes are connected to each other irrespective of whether there exists a path between them or not. For our case, the path is a unit step that can be taken by the agent in any of the four directions. The input feature at each node $x_i$ is equal to the $(1 \times n)$ dimensional grid element $p_{uvl}$ discussed earlier, with number of input features $D=n$. The expected output at each node is a one-hot encoding class wise representation for the following 4 possibilities ($F=4$):
\textbf{Unknown}(e.g.,- beyond a visible wall)- encoded as $[1,0,0,0]$
\textbf{Navigable space}- encoded as $[0,1,0,0]$,
\textbf{Target}- encoded as $[0,0,1,0]$ and
\textbf{Obstacle}- encoded as $[0,0,0,1]$,

\textbf{Supervised training-}

Algorithm ~\ref{graphconv} shows the details of forward computation and update of neural network parameters $f$ for our proposed GCN. Let $L$ be the total number of graph convolution layers, $f^l$ denote neural network parameters at a particular layer of GCN, $V$ and $E$ denotes the collection of vertices and edges in the graph structure. First the node level features $x_i$ are aggregated (summed together) over its connecting neighbors, after which a global function for that layer $f^l$ is applied to get global feature of the graph for that layer $l$ as $Z^l$. The final global feature $\tilde{Z}$ is compared to the problem specific global feature labels (see paragraph above) using negative log likelihood loss function (NLLLoss), later all the parameters in all the layers $f$ are updated using backpropagation with Stochastic Gradient Descent (SGD).

\begin{algorithm}

\DontPrintSemicolon
  \KwInput{Graph $G = (X,V,E)$, update parameters $f$}
  \KwOutput{ $G = (\tilde{Z},V,E)$, new parameter $\tilde{f}$ }
  \KwData{Target output graph $G = (Z,V,E)$}
  \For{$l \in L$} 
  {
        \For{$i \in \{1,\dots,|V|\}$}
      {
      $x^l_i \gets x^{l-1}_i+\sum_{(i,j)\in E} x^{l-1}_j$
      }
      $X^l \gets \{x^l_1,\dots,x^l_{|V|}\}$
      
      \tcp{apply multilayer perceptron}
      $Z^l \gets f^l(X^l)$
  }
  $\tilde{Z} \gets Z^L$
  
  $\mathcal{L} = NLLLoss(\tilde{Z},Z)$
  
  $\tilde{f} \gets SGD(f,\mathcal{L})$

\caption{GCN-computation + parameter update}
\label{graphconv}
\end{algorithm}

\section{Contextual understanding of natural language instructions}
\begin{figure}
  \centering
  \includegraphics[width=0.50\textwidth]{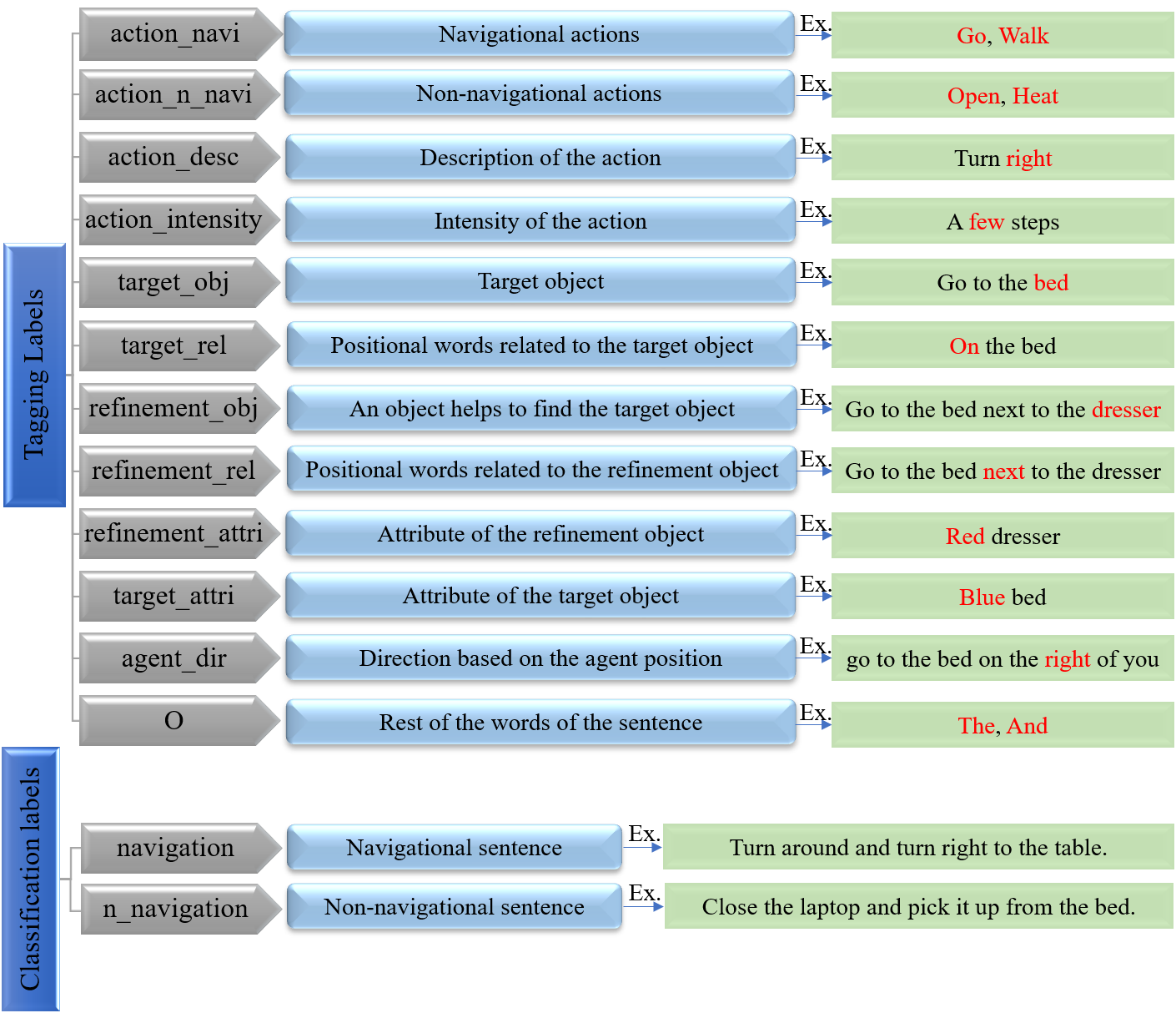}
  \caption{\textit{Defined labels for intent classification and slot filling tasks for Alfred dataset.  }}
  \label{fig:slot_filling}
\end{figure}

This module, extracts relevant interpretations of the input instructions that could be pieced together with information from the other modules. We focus on extracting two types of interpretations for the instructions simultaneously.
First, each sentence is classified as either a navigation task or a non-navigation task- $y^i$ (known as \emph{intent classification}, with labels - navigation and not-navigation).
Second, each word in these instructions is classified into a semantic category (slot labels), which indicate the role of a word and its semantic relations to the other words in the sentence (this is a sequence labeling task known as \emph{slot filling}). Thus, the sequence of words in the input $x = \{x_1,\dots,x_T\}$ is labeled with the slot label sequence $y^s = \{y^s_1,\dots,y^s_T\}$ in the output. In the proposed language understanding module, two labels for intent classification and 12 labels for slot filling (word-level semantics) are considered (see figure~\ref{fig:slot_filling}). To simultaneously classify the intent and the label sequence of an instruction, we leverage the recently proposed Bidirectional Encoder Representations from Transformers (BERT)~\cite{vaswani2017attention} approach. This model is used as a source model for transfer learning purposes~\cite{chen2019bert}. 

The input representation to BERT is a concatenation of WordPiece embeddings~\cite{wu2016google}, positional, and segment embeddings. With a special classification embedding ([CLS]) inserted as the first token and a special token ([SEP]) as the final token, BERT takes the token sentence $x = (x_1,\dots,x_T)$ as input and output hidden states $H=(h_1,\dots,h_T)$. Based on hidden state of [CLS], the intent of a sentence is predicted as $y^i = \softmax(W^ih_1 + b^i)$. The final hidden states of other tokens $h_2,\dots,h_T$ can be fed into a softmax layer to assign slot labels to words. Each tokenized input word is fed to the WordPiece tokenizer and the hidden state for the first sub-token (call it $h_n$ corresponding to $x_n$) is input to a softmax classifier to get the label of the $n^{th}$ word. Let $N$ be number of tokens, then:
\begin{equation}
    y^s_n = \softmax(W^sh_n + b^s), n \in 1,\dots,N
\end{equation}
The following objective models the joint task of intent classification and slot filling
\begin{equation}
    p(y^i,y^s|x) = p(y^i|x)\prod_{n=1}^N p(y_n^s|x)
\end{equation}
Finally we maximize the conditional probability $p(y^i,y^s|x)$, using cross-entropy loss to train/fine-tune the model.

\begin{figure*}
  \centering
  \includegraphics[width=0.60\textwidth]{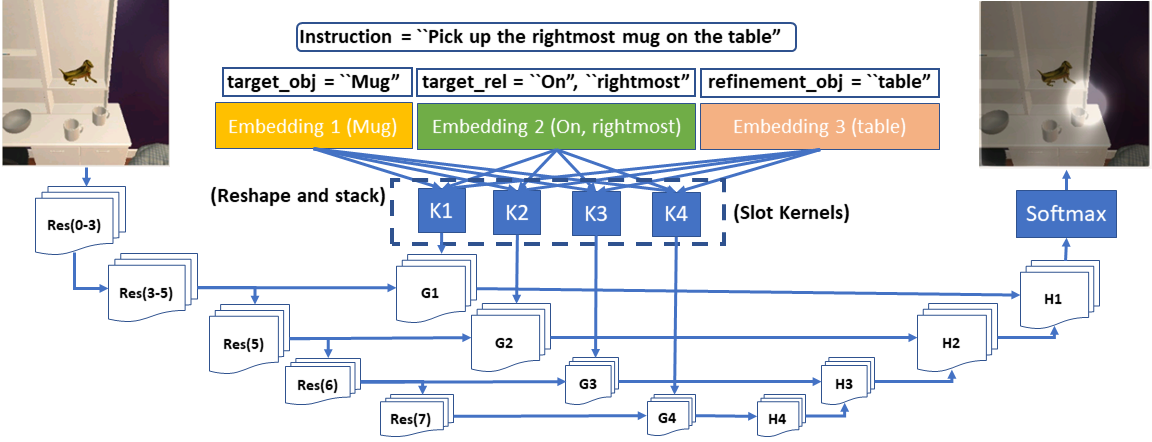}
  \caption{\textit{Cross modal grounding framework adapted from LingUNet. Res(i-j) denotes pretrained layers of ResNet starting from layer number i to layer number j. Output of softmax is intersected with output of UNet based class segmentation map to arrive at final map.   }}
  \label{cross_modal_framework}
\end{figure*}

\section{Vision language grounding}
A major challenge in vision language navigation is making the connection between language instructions and observed visual inputs. A technique that segments images based on key words in a natural language expression is known as referring expression image segmentation. 
We adapt the LingUNet learning framework~\cite{misra2018mapping} for predicting a probability distribution over input pixels of RGB image. We take a slight departure from the original formulation for presenting instruction encodings to a UNet architecture~\cite{ronneberger2015u}.

As shown in figure~\ref{cross_modal_framework}, we pass the text instruction through the language understanding module which assigns slot labels to each word in the sentence according to the labels proposed in this paper (as shown in figure~\ref{fig:slot_filling}). The words that get assigned labels are converted to embedding vectors through a learnable embedding layer. The embedding obtained for each slot label are reshaped into a kernel and stacked together as a set of filters. Each of these stack of filters $(K1,K2,K3,K4)$ are called slot kernels and are convolved over feature representations obtained from forward pass through the pretrained ResNet~\cite{he2016deep} layers. In the example shown in figure~\ref{cross_modal_framework}, an image is extracted from the ALFRED dataset with Ai2Thor simulator. The input image gets a forward pass through the pretrained ResNet layers- $Res(i-j)$, where $i$, and $j$ are the layer numbers, and $i-j$ implies the network formed by the layers $i$ through $j$. Before upsampling through convolution filters $H1$ to $H4$, they are convolved with slot kernels to obtain $G1$ to $G4$. The final feature map, after application of Softmax operation, assigns a probability distribution which highlights regions having a high probability of being the object mentioned in the text instruction. This prediction is further combined with instance segmentation to get the exact object.

\section{Results and performance comparison}
In this paper, we use the recently proposed ALFRED data set~\cite{shridhar2020alfred} built upon the Ai2Thor simulator~\cite{kolve2017ai2} to demonstrate the efficacy of our proposed \textbf{MoViLan} framework. Contrary to most other simulators, it features extremely long execution trajectories (some requiring upto 100 low level actions) and an immense object interaction diversity featuring compositional tasks such as in figure~\ref{fig:task_ex}, thereby allowing us to demonstrate performance in significantly harder and closer to real life scenarios. 

\textbf{Tasks description - }
Scenarios are divided into collection of compositional tasks (language instructions with expert demonstrated sequence of low level actions -producing RGB images) over 120 different rooms including kitchens, living rooms, bedrooms and bathrooms (30 rooms each). Tasks demonstrated in ALFRED can be classified into 7 different types - Pick and Place, Stack and Place, Pick two and Place, Clean and Place, Heat and place, cool and place and examine in light. More closely, these tasks are a combination of 8 fundamental high level ``sub-goals"- GoTo (navigate to some place), Pickup (pick up small objects like a pen), Put (place objects that have been picked up in a certain receptacle), Cool (cool an object), Heat (heat up an object), Clean (remove objects on a surface), Slice (cut an object) and Toggle (turn appliances on and off). 

\textbf{Training, test and parameters - }
    \textbf{Training dataset} - We selected a subset of the ALFRED dataset for instructions over living rooms (Ai2Thor room number 201-230) and bedrooms (301-330). Dataset authors proposed train/test split which has a much lesser diversity of test unseen rooms compared to train. This could provide extra advantage to our mapping module which is trained independently for just localizing in rooms. Therefore we propose a train/test split where all training data (including language instructions) is extracted from rooms 311-330 and 211-230 leaving aside rooms 301-310 and 201-210 for testing. We compare performance with that of sequence-to-sequence methods presented in~\cite{shridhar2020alfred}, referred to as the \emph{baseline} for the remainder of the paper. While the \emph{baseline} is tested on the same unseen rooms, it is allowed to train on the remaining data set (40 rooms). The 20 test rooms have roughly 6000 natural language instructions. 
    \textbf{Language Understanding} - From our training data, we hand label 1000 language instructions using our proposed labeling scheme and finetune on a pretrained BERT model for slot filing and intent detection tasks. Training parameters are the same as in~\cite{JointBERT}.
    \textbf{Mapping} - On train rooms we scan panorama images at each navigable position in the room, approximate projection maps, and simultaneously form ground truth occupancy/collision maps (upon placement of an agent at each point in the room). The training of the graph convolution is to input approximate projections and estimate the corresponding collision grid map around the agent. (navigable space, obstacle, target, unmappable (beyond wall)). Training parameters are same as in~\cite{Mapping}.
    \textbf{Vision language grounding} - BERT provides semantic slot labels for each word of the sentence. Therefore during test, instead of natural instructions, structured words can be provided as in~\ref{cross_modal_framework}. Therefore to train LingUNet with our proposed input restructuring, human level natural language is not required. We can automatically generate training data as follows. Images (300x300 pixels) are extracted from each position of training rooms. From the ground truth segmentation of the image upto 3 objects can be selected randomly and their contours and center points are extracted. Their names are collected from the metadata of the simulation. Using a random ordering now each object can be referred to the other based on relative locations of their center points. Training parameters are same as in~\cite{VisionLanguage}. No training is required for disambiguation and planing modules.



\textbf{Agent observation space -}
Since Ai2Thor does not support panorama images during test time our agent acquires panorama images just as in the top left corner of figure~\ref{fig:1framework}. Instance segmentation (middle) is estimated using UNet, and depth estimation (bottom) is used as provided by the simulator, which can also be replaced by off the shelf monocular depth estimation frameworks. The agent maintains a persistent map throughout execution of the provided instructions updated every time it takes an action using our proposed graph convolution filtering approach. On encountering collisions, it is registered into the persistent map and is recognized while taking subsequent actions.

\textbf{Agent Action Space - }
From the simulator we chose a multitude of low level actions as listed in-~\cite{Allen2020}. Total of 13 discreet actions available each time step. For example, movement actions like - MoveAhead, RotateRight, LookUp, LookDown, and interaction actions like - Pickup, Put, Open, Close, ToggleOn, etc. Interaction actions require to either provide an interaction mask over input RGB image, or provide the complementary interacting object on which to act. While the \emph{baseline} chose the former approach, we adopt the latter approach.

\begin{figure*}
    \centering
    \includegraphics[width=.65\linewidth]{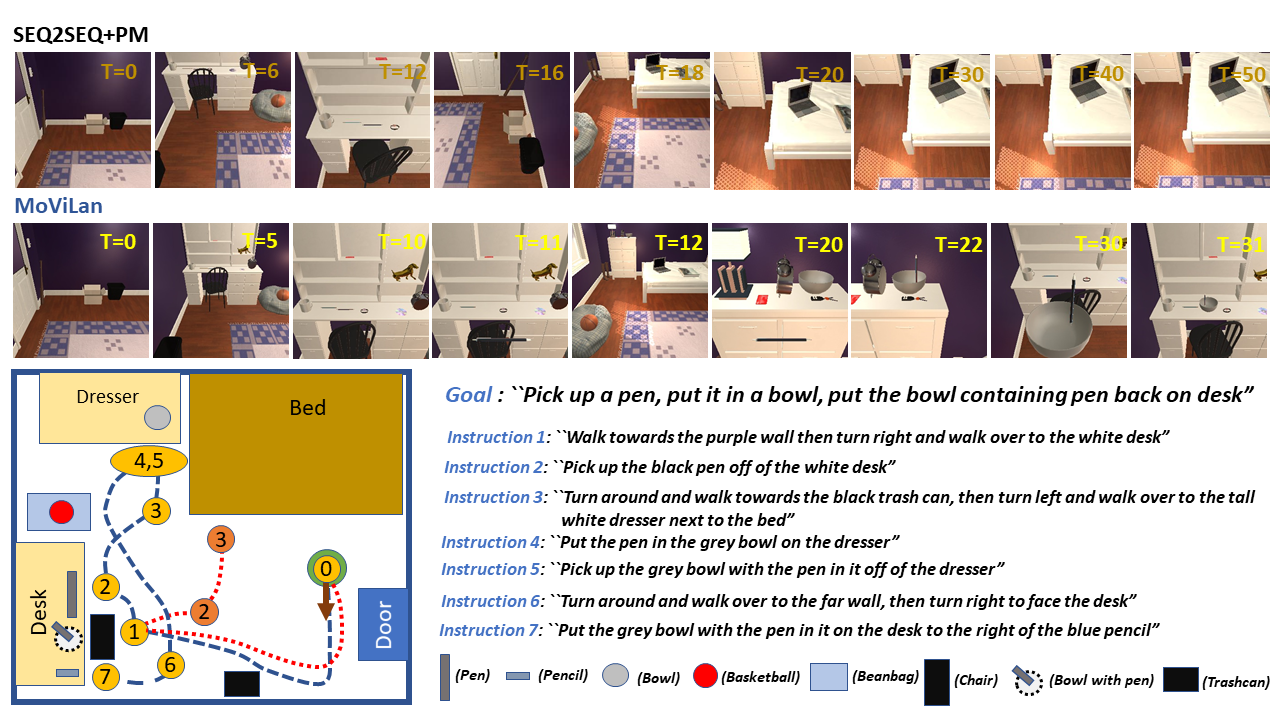}
    \caption{\textit{Comparison of trajectories for the \emph{baseline} and our methods; agent shown by green circle, with starting orientation shown by brown arrow; relevant subgoals corresponding to instructions are shown in yellow (ours) and orange (author) colored circles; trajectory for our method shown by dotted blue line (success) and that for \emph{baseline} shown by dotted red line (failure).}}
    \label{fig:traj_compare}
\end{figure*}

\textbf{Metrics - }
We use the same metrics used for evaluating the \emph{baseline} in~\cite{shridhar2020alfred}:
    \textbf{Task Success} is a binary metric tracked for each task. Value is set to 1 if all the ``sub-goals" that constitute execution of a complete task specified by the entire language instruction have been fulfilled. Otherwise this is set to 0. ``Success-rate" of an algorithm is then calculated by taking the average of all task success values over all the tasks.
    \textbf{Goal-Condition Success} is the ratio of ``sub-goals" completed successfully out of all the goals in the task. The sub-goal success of an algorithm is similarly an average over all tasks.
    \textbf{Path weighted success rate and goal-condition success} are computed as multiplying the task success and the goal condition success rates respectively by the ratio $\frac{r}{R}$, where $R$ is the maximum of number of required actions (expert demonstrated) and number of low level actions executed by the agent, and $r$ is the number of agent actions. 
    Finally, \textbf{Sub-Goal success rate} is the ratio between the number of times a sub-goal is achieved and the total number of sub-goals.

\subsection{Baseline methods}
\textbf{Seq-to-Seq -}
The first \emph{baseline} technique is a sequence-to-sequence (SEQ2SEQ) CNN-LSTM model. In this architecture, a Convolutional Neural Network (CNN) encodes the visual input, a bidirectional-LSTM generates a representation of the language input, and a decoder LSTM infers a sequence of low-level actions while attending over the encoded language. 
In another variant of the \emph{baseline} approach, called SEQ2SEQ+PM, two auxiliary losses are included that use additional temporal information to make sequence to sequence learning more tractable on long trajectory tasks. Please refer to~\cite{shridhar2020alfred} for further details. 
\subsection{Benchmark comparison}
Tables~\ref{tab:my-table1} and~\ref{tab:my-table2} provide performance comparison for complete tasks and modular sub tasks respectively. In ``MoViLan + PerfectMap", ground truth BEV maps are provided to the agent as an ablation study for removal of our mapping module. In supplementary we conversely compare performance gains provided by our mapping module compared to pure geometric projection for targeted navigation tasks. Our framework demonstrates superior performance compared to the \emph{baseline} algorithms on complete tasks. For sub-goal tasks, our framework has significantly higher path weighted success rates for ``GoTo" (language instructions requiring pure navigation), and ``PickUp". Note that with PerfectMap, task success rate increases significantly without any considerable increase in sub-goal success rate for ``PickUp",``Put" and ``Toggle". This is because without good positioning, the agent would fail on many manipulation tasks. Anecdotally, figure~\ref{fig:traj_compare} demonstrates a failure case (top row) for \emph{baseline} and a corresponding success case (bottom row) for our framework. It is observed that the \emph{baseline} exhibits a behavior of ``memorizing" trajectories from training data. While it leads to trajectory lengths close to expert demonstration in a small number of success cases, it also succumbs to failure in unseen rooms and for unseen object interactions. This conjecture is further supported by the fact that we observe \emph{baseline} success rate as high as 25\% on the training set. Authors in~\cite{shridhar2020alfred} have also highlighted the higher success rates for \emph{baseline} on sub-goal tasks. This is an indicator that models using modularity or hierarchy would be favorable as we find in our case.
\begin{table}[htbp]
\centering
\begin{tabular}{|l|l|l|l|l|}
\hline
\textbf{Test (unseen)(\%)} & \multicolumn{2}{l|}{Task} & \multicolumn{2}{l|}{Goal-Cond} \\ \hline
SEQ2SEQ & 2 & (1) & 2 & (1) \\ \hline
SEQ2SEQ+PM & 3 & (2) & 4 & (3) \\ \hline
MoViLan & 37 & (18) & 45 & (22) \\ \hline
MoViLan + PerfectMap & 60 & (35) & 72 & (40) \\ \hline
\end{tabular}
\caption{Task and Goal-Condition success percentages (rounded to nearest integer percentage). Path length weighted metrics are provided in parenthesis.}
\label{tab:my-table1}
\end{table}

\begin{table}[htbp]
\centering
\begin{tabular}{|l|l|l|l|l|}
\hline
\textbf{Test (unseen)(\%)} & GoTo & Pickup & Put & Toggle \\ \hline
SEQ2SEQ & 30 & 26 & 41 & 31 \\ \hline
SEQ2SEQ+PM & 31 & 29 & 43 & 30 \\ \hline
MoViLan & 45 & 50 & 49 & 60 \\ \hline
MoViLan + PerfectMap & 70 & 51 & 53 & 62 \\ \hline
\end{tabular}
\caption{Sub-Goal success rates (path length weighted) (rounded to nearest integer percentage) on unseen test data}
\label{tab:my-table2}
\end{table}

\begin{figure*}
    \centering
    \includegraphics[width=0.7\linewidth]{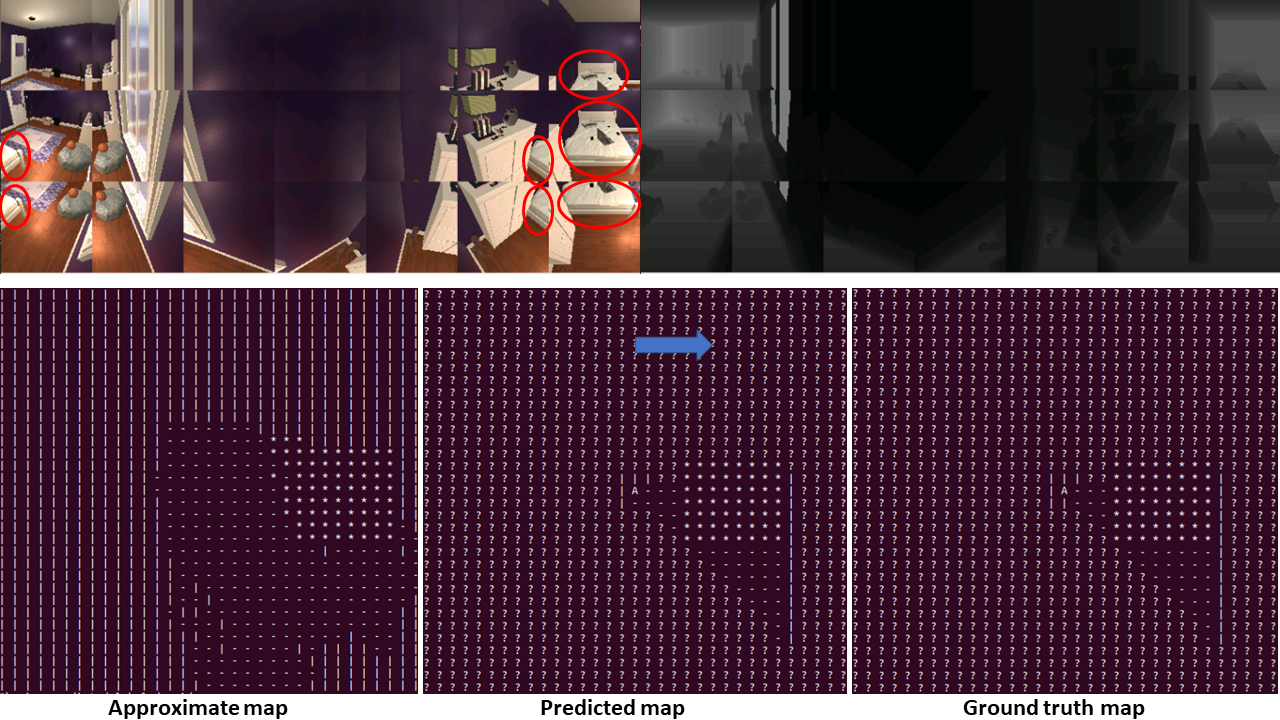}
    \caption{\textit{Captured panorama image by the agent (top left), captured depth normal by the agent (top right), and bottom row showing projected maps. Target object ``Bed” is selected in red circle. Bottom row leftmost shows the approximated projection map. Here each grid location in the map is a vector of 4 representing probabilities of that location being a target (marked ‘*’), navigable space (marked ‘-’), obstacle (marked ‘I’) and unknown (because of obstruction, marked ‘?’). For concise representation, only the argmax value at each grid node is represented by the corresponding marker, meaning if a location has star, it has the highest probability for being the target object. For purposes of planning, unknown is treated the same as obstacle. Input approximate projection (bottom row left) to our proposed Graph convolution filtering algorithm gives the predicted map (bottom row mid). Agent ego centric north is shown by the blue arrow and the ego center (perceived location of self) is shown by the letter ‘A’.
 }}
    \label{map1}
\end{figure*}

\begin{figure*}
    \centering
    \includegraphics[width=0.7\linewidth]{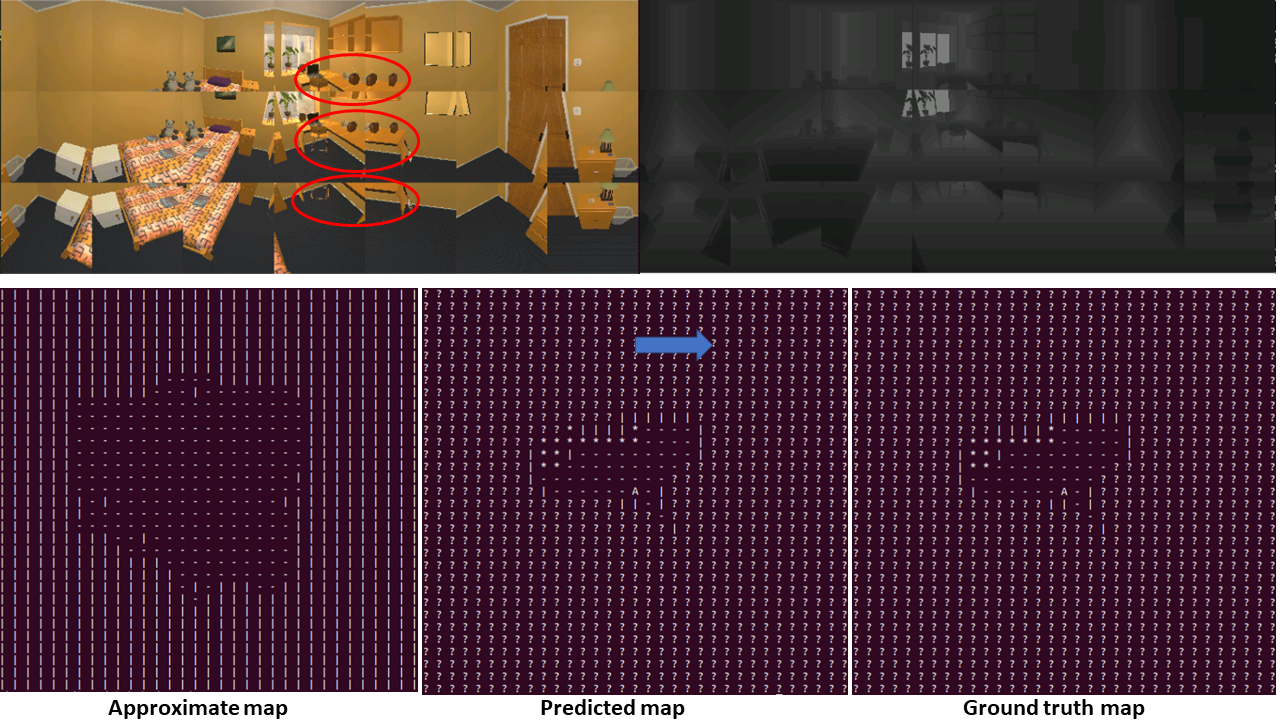}
    \caption{\textit{Captured panorama image by the agent (top left), captured depth normal by the agent (top right), and bottom row showing projected maps. Target object ``Desk (L shaped)” is selected in red circle. Bottom row leftmost shows the approximated projection map. Here each grid location in the map is a vector of 4 representing probabilities of that location being a target (marked ‘*’), navigable space (marked ‘-’), obstacle (marked ‘I’) and unknown (because of obstruction, marked ‘?’). For concise representation, only the argmax value at each grid node is represented by the corresponding marker, meaning if a location has star, it has the highest probability for being the target object. For purposes of planning, unknown is treated the same as obstacle. Input approximate projection (bottom row left) to our proposed Graph convolution filtering algorithm gives the predicted map (bottom row mid). Agent ego centric north is shown by the blue arrow and the ego center (perceived location of self) is shown by the letter ‘A’.
. }}
    \label{map2}
\end{figure*}

\begin{figure*}
    \centering
    \includegraphics[width=0.7\linewidth]{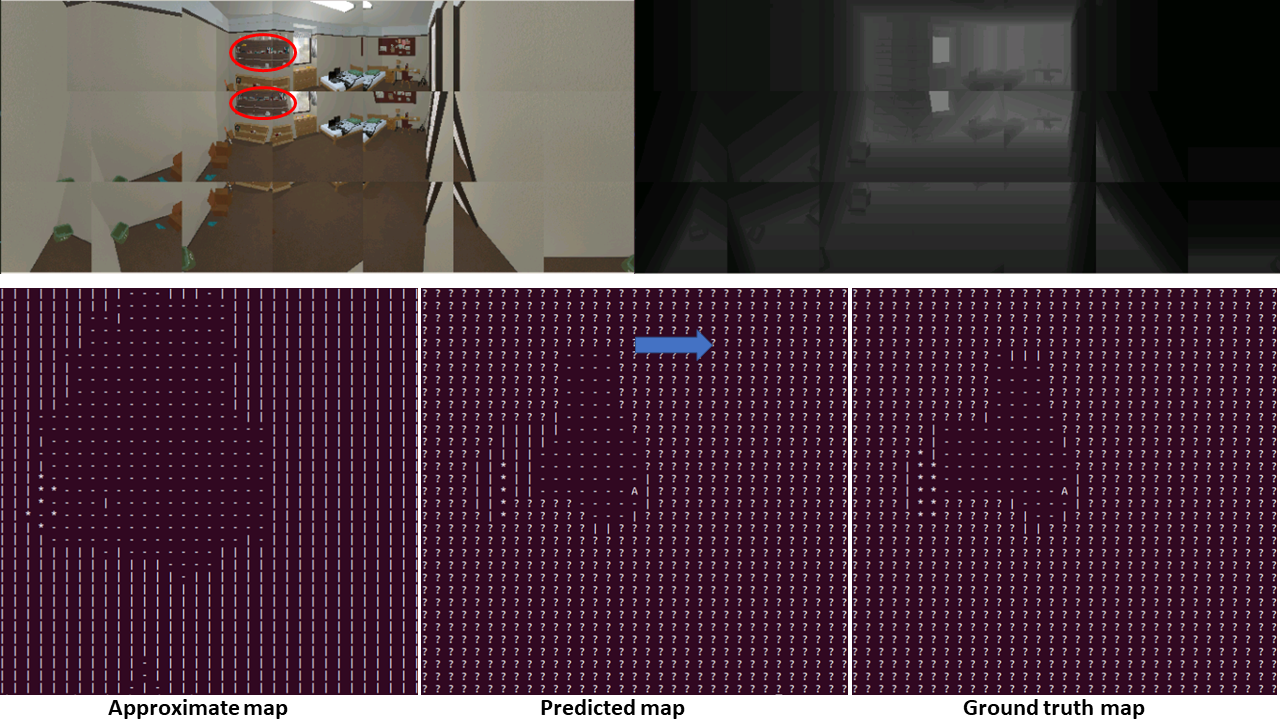}
    \caption{\textit{Captured panorama image by the agent (top left), captured depth normal by the agent (top right), and bottom row showing projected maps. Target object ``Shelf” is selected in red circle. Bottom row leftmost shows the approximated projection map. Here each grid location in the map is a vector of 4 representing probabilities of that location being a target (marked ‘*’), navigable space (marked ‘-’), obstacle (marked ‘I’) and unknown (because of obstruction, marked ‘?’). For concise representation, only the argmax value at each grid node is represented by the corresponding marker, meaning if a location has star, it has the highest probability for being the target object. For purposes of planning, unknown is treated the same as obstacle. Input approximate projection (bottom row left) to our proposed Graph convolution filtering algorithm gives the predicted map (bottom row mid). Agent ego centric north is shown by the blue arrow and the ego center (perceived location of self) is shown by the letter ‘A’.
 }}
    \label{map3}
\end{figure*}

\begin{figure*}
    \centering
    \includegraphics[width=0.7\linewidth]{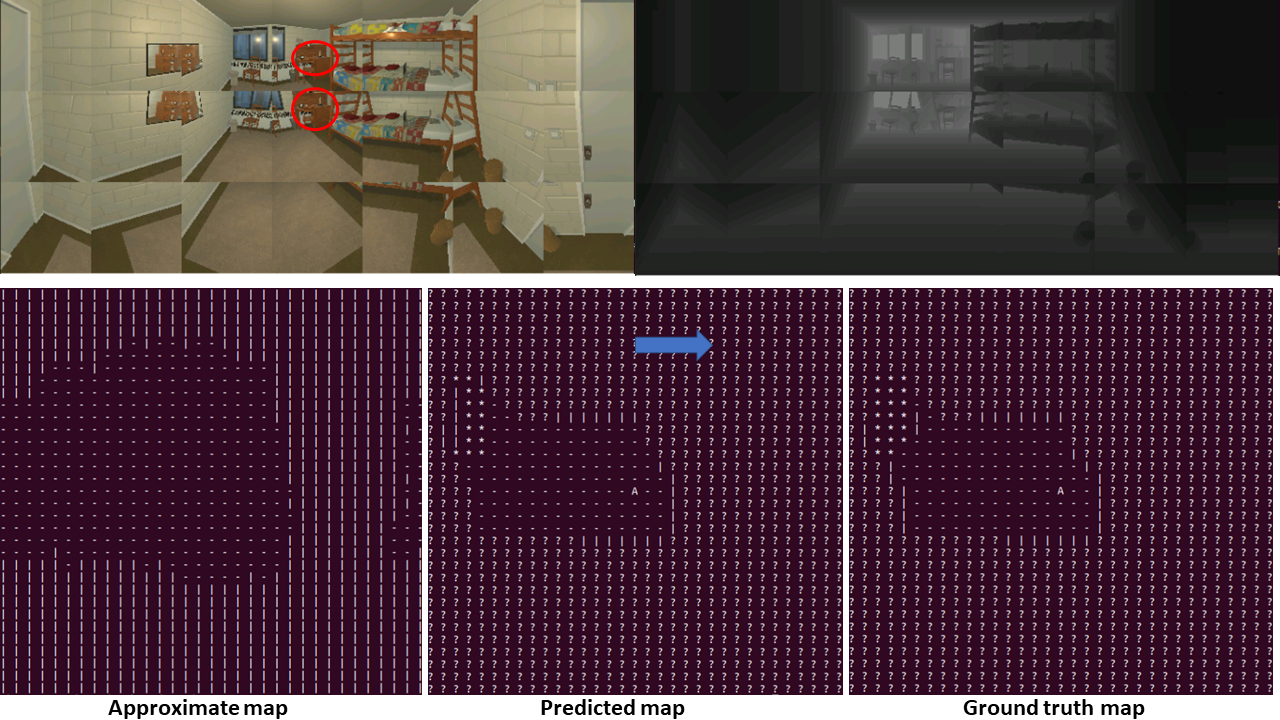}
    \caption{\textit{Captured panorama image by the agent (top left), captured depth normal by the agent (top right), and bottom row showing projected maps. Target object ``Dresser” is selected in red circle. Bottom row leftmost shows the approximated projection map. Here each grid location in the map is a vector of 4 representing probabilities of that location being a target (marked ‘*’), navigable space (marked ‘-’), obstacle (marked ‘I’) and unknown (because of obstruction, marked ‘?’). For concise representation, only the argmax value at each grid node is represented by the corresponding marker, meaning if a location has star, it has the highest probability for being the target object. For purposes of planning, unknown is treated the same as obstacle. Input approximate projection (bottom row left) to our proposed Graph convolution filtering algorithm gives the predicted map (bottom row mid). Agent ego centric north is shown by the blue arrow and the ego center (perceived location of self) is shown by the letter ‘A’.
 }}
    \label{map4}
\end{figure*}

\begin{figure*}
    \centering
    \includegraphics[width=1.0\linewidth]{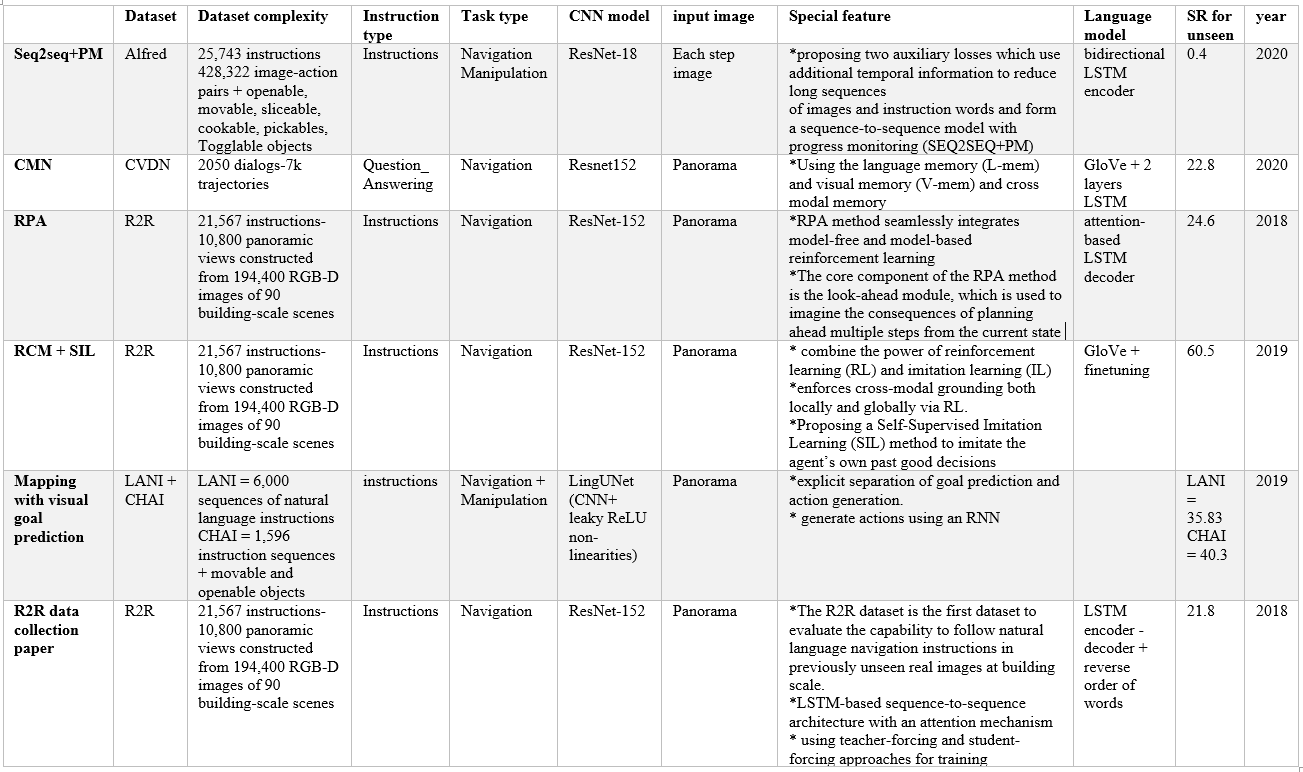}
    \caption{\textit{Review of major techniques in vision language instruction following. }}
    \label{reviewtab}
\end{figure*}

\section{Supplementary Material}
This section covers detailed review of related works in this field as well as a broad range of ablation studies and detailed explanations for all the five different modules of our framework.

\subsection{Detailed related work}
Here we present a detailed review of all approaches used in vision language instruction following, divided into domains based on mapping, language understanding and end to end approaches. Figure~\ref{reviewtab} also provides a summary of end to end approaches along with various datasets.
\\
\\
\\
\\
\\
\\
\\

\subsubsection{Mapping based approaches}
Central to many robotic tasks is to learn and maintain a metric map to rely on for navigation. 
Early work by the authors~\cite{smith1986representation,leonard1991simultaneous}, laid the foundations for the probabilistic formulation of building globally consistent maps, which is now widely known as Simultaneous Localization and Mapping (SLAM). Over potentially long explorations by the robot, data from a variety of sensors are integrated~\cite{bongard2008probabilistic,fuentes2015visual}. SLAM can be broadly divided into filter based and graph based. In the former, the temporal aspect of consecutive measurements is emphasized, whereas in the latter spatial aspects such as robot poses and visibility of landmarks is important~\cite{grisetti2010tutorial}. In recent works for vision language navigation~\cite{anderson2019chasing}, authors use a metric map as memory architecture for navigating agents. In topological maps, the graph convention can be used to represent environments with nodes representing free space with edges between the nodes connecting the free spaces by traversable paths. While metric maps focus on the geometry of the surrounding space of the agent, topological maps are more geared towards the connection between robot poses and trajectories in the environment, which often provides simpler solutions to route based navigation~\cite{werner2000modelling}. 
Introduction of semantics in graph based maps can provide powerful solutions to encode relative locations of objects of interest in persistent memory~\cite{pronobis2012large,lang2014definition} and is useful when a higher level understanding of the surrounding is necessary. Deep learning approaches have also been proposed in this context, as in neural SLAM~\cite{zhang2017neural} that tries to mimic the SLAM procedures into soft attention based addressing of external memory architectures. Authors in~\cite{chaplot2020learning} also extend upon these approaches in a modular and hierarchical fashion. Several studies have also used differentiable and trainable planners for navigation~\cite{tamar2016value,khan2017end,lee2018gated}. 
There is another branch of work that focuses on accurately navigating to a location once a map has already been provided. 
They mainly focus on ``targetted-navigation" and dont deal with the complexities of natural language. A major challenge they try to solve is trying to separate uncertain part from the certain part (ex- for a given room layout where you already know the position of big furnitures, where might small things like keys, remote control, etc be)~\cite{sunderhauf2019keys}. This is an important concept when instructed language involve interaction with small objects in certain parts of an environment (not observable from far away, or may need to perform some actions to observe such as open a drawer), not just navigate to a place. 
Although detailed topological maps may be provided, localization in dynamic settings may not be trivial. Authors in~\cite{chen2019behavioral} explore a behavioral approach to navigation using graph convolution network over a topological map.
In this regard, our contribution is mainly constructing dense semantic topological maps from panorama images, specialized for cluttered indoor environments.

\subsubsection{Approaches based on language understanding}
Earliest works on interpreting natural language commands for navigation have used statistical machine translation methods for mapping instructions to formal language~\cite{matuszek2010following,chen2011learning}. In~\cite{matuszek2010following}, the authors had used general purpose semantic parser learner (WASP)~\cite{wong2006learning} in order to learn semantic parser and constrain it with the physical limitations imposed by the environment. 
With the recent success of neural networks in natural language processing~\cite{goodfellow2016deep}, several techniques have been developed for interpreting user commands framed as a sequence prediction problem~\cite{mei2016listen,anderson2018vision}. Such machine level translation of natural language instructions has also been explored in the context of automatic question answering~\cite{xiong2016dynamic,seo2016bidirectional}. Following up, recently authors in~\cite{zang2018translating} uses attention mechanisms to learn alignment of language instructions with given topological map and output high level behaviors. 

In contrast, we adopt a completely unique approach that associates semantics to each word of the language instruction using a semantic slot filling model. We find that the semantics we define are easy to learn and general enough to encompass various kinds of household instructions. 
This allows to significantly reduce the complexity of input language instructions, allowing simple search techniques to execute these instructions on a semantic topological map (which we also learn using our new proposed technique- see mapping section).
We will release the training data for this semantic slot filling model (that we use to fine-tune BERT). We believe that this will help researchers to develop advanced techniques on top of our proposed instruction simplification mechanism.

\subsubsection{End to end learning approaches}
Several recent deep learning approaches propose to learn a mapping directly from inputs to actions, whether structured observations are provided~\cite{mei2015talk,suhr2018situated} or the agent deals with raw visual observations~\cite{misra2017mapping,xiong2018scheduled}.
Proponents of this approach argue about the simplicity of these models in outperforming a combination of models that require engineering hand crafted representations. 
Cross modal grounding of language instructions to visual observations is often used in several works, via e.g., reinforcement learning~\cite{wang2018look,wang2019reinforced}, autoencoder architectures that impose a language instructions-based heat map on the visual observations (using U-net architectures~\cite{misra2018mapping}, attention mechanisms~\cite{zhu2020vision}, or implementation of non linear differentiable filters~\cite{anderson2019chasing}). However, as we show later in results, going end to end may not be best for generalizing to perform compositional tasks in unseen environments.

However, these frameworks often may not combine depth sensing and semantic segmentation which are extremely valuable sources of information and can be easily transferred from pre-trained frameworks. 
Also keeping depth sensing and segmentation modules separate helps to incorporate knowledge about new unseen object categories into the framework in an easier fashion than in end to end models. 
Several past research have tried to learn input language and expert action demonstrations into a joint feature representation using such reinforcement/seq-to-seq learning techniques, but judging from the nature of language instructions (eg- R2R dataset) it can be observed that most often agents only require to identify a target object on the map and navigate to it. This situation can become much more complicated with compositional instructions common in household robotics such as-``open the microwave, put the coffee, and wait for 5 seconds''. Although the number of expected actions from the language is not immense, the complexity of this joint mapping from vision and language to action may increase drastically, even more with longer sentences. This might make existing sequence to sequence or reinforcement learning techniques much harder to learn. 
For this reason, we advocate new individual mapping and language understanding modules and combine them using extracted semantic priors. 

\subsection{Approximate Birds Eye View (BEV) projection mapping}
Given an RGB image $x_r \in \mathcal{R}^{h\times w\times 3}$, a corresponding segmented image $x_s \in \mathcal{R}^{h\times w\times n}$ (each pixel stores a one hot encoding denoting one out of $n$ possible object classes), and a corresponding depth normal image $x_d \in \mathcal{R}^{h\times w\times 1}$, our goal is to obtain a projection $p\in \mathcal{R}^{s\times s\times n}$. Here, $h$ is the height of the RGB image, $w$ is the width, $n$ is the number of object classes of interest, and $s$ is the size of the spatial neighborhood around the agent where we are projecting to obtain a map. Each grid of the $s\times s$ neighborhood contains a $1 \times n$ dimensional feature vector denoting one out of $n$ possible object classes occupying the grid (including navigable space). Standard image segmentation frameworks and monocular depth estimation frameworks can be used to provide $x_s$ and $x_d$ from $x_r$ with relative ease. Each pixel value of depth image can be multiplied with pixel indices, camera focal length and calibration matrix to obtain $x,y$ and $z$ distance of the pixel with respect to the agent, which is then rescaled and normalized to lie in any of the $s \times s$ grid around the agent (detail equations in next section). Therefore, pixel $i,j$ in $x_r,x_s$ and $x_d$ get assigned to the grid location $u(i,j)$ and $v(i,j)$ around the agent in $p\in \mathcal{R}^{s\times s\times n}$. Let $(u,v)$ be the shorthand for each grid location which now stores a $(1\times n)$ vector. Here $n$ is the number of objects of interest. Now, let each element position $l$ in the $(1\times n)$ vector store 1 if the grid location was mapped from the contributing pixel in RGB image which belonged to class $n$, and 0 otherwise. Therefore, $p_{uvl}$ can either store 1 or 0. Furthermore, as we are using panorama images, a value of $p_{uvl}$ is obtained for each discrete rotation of the agent made to complete the image (in many applications, 360 field of vision is not possible, so agent has to rotate in steps). Values $p_{uvl}$ obtained for each rotation is multiplied with rotation matrix $R(\theta)$ and averaged to obtain the final approximate projection map $p$.

\subsubsection{Projection mapping technique}
This part is an extension to the main text and describes the equations used for obtaining the approximate projection map.

Ground projection in general is not a one-to-one mapping. That is, 0, 1 or many more elements of the ($1\times n$) image features from $x_s$ is possible to be ground projected onto any element of $p\in \mathcal{R}^{s\times s\times n}$. For this reason, uncertainty and noise removal is very important in mapping which we primarily achieve through our proposed graph convolution framework described in the next subsection. To resolve ambiguities, some frameworks either take the maximum of the competing elements~\cite{henriques2018mapnet,qi2017pointnet}, or some frameworks also take the aggregate of all the elements. We take a different route and allow competing elements to add up along the respective dimensions of the $1\times n$ vector for all different competing elements of different object categories. This noisy feature vector is filtered later by our graph convolution algorithm to contain the most probable object in that grid.

Let $K \in \mathcal{R}^{3 \times 3}$ be the agent camera calibration matrix. Let $d_{ij}$ denote the $(i,j)$ pixel in the depth normal image $x_d$. Let $p_x(i,j)$ denote the $x$ distance (relative to agent) of the $(i,j)^{th}$ pixel in the RGB image, accordingly, $p_y(i,j)$ and $p_z(i,j)$. They can be obtained using the following formula. 

\begin{table*}[htbp]
\centering
\begin{tabular}{|l|l|l|l|l|}
\hline
                            & Navigable space           & Big Targets          & Medium Target       & Small Targets       \\ \hline
Approx. projection          & 65.1 $\pm$ 10 & 62.3 $\pm$ 7 & 60.2 $\pm$ 5 & 55.3 $\pm$ 5 \\ \hline
Graph convolution filtering & 95.5 $\pm$ 2  & 93.2 $\pm$ 2 & 91.4 $\pm$ 3 & 89.6 $\pm$ 5 \\ \hline
\end{tabular}
\caption{Table showing comparison of performance improvement in predicting the class of each node in grid projection map when using Graph convolution filtering over approximate projection map as compared to only approximate projection map. Values are shown in mean accuracy in percentage of correctly assigned grids followed by standard deviation values. Four major categories are investigated (along columns)- navigable space, identifying grids containing big targets like Bed, identifying medium sized targets like Desk, dresser, and identifying small targets like SideTable, Stool.}
\label{mapping-table}
\end{table*}

\begin{gather}
 \begin{bmatrix} p_x(i,j) \\ p_y(i,j) \\ p_z(i,j)  \end{bmatrix}
 =
 K 
  \begin{bmatrix}
  i \\ j \\ f 
  \end{bmatrix}
    d_{ij}, i \in \{1,\dots,h\}, j \in \{1,\dots,w\}
\end{gather}
Let $V$ denote the maximum range of vision of the depth camera or depth estimation method, $D$ denote the maximum pixel value of $x_d$, and $r$ denote the edge length of each small grid in the projection map (or resolution). A contributing element of the projection map $p\in \mathcal{R}^{s\times s\times n}$ can be obtained from the pixel $(i,j)$ as follows:
\begin{align}
    u(i,j) = \frac{V}{D\times r}p_x(i,j) - \frac{s+1}{2}
    \\
    v(i,j) = \frac{V}{D\times r}p_z(i,j) - \frac{s+1}{2}
\end{align}
Each grid in the projection map $p$ can now be indexed by the three values $u$, $v$, and $l$, where we are defining $l$ as the index of the $(1\times n)$ vector denoting a certain object present at the grid location $(u,v)$. We let $class(u(i,j)) \in \{1,\dots, n \}$ be the function that returns the object category of the grid out of the $n$ possible classes. This can be directly obtained from $x_s$. We can now use the following simple rule to assign the value of $p_{uvl}$. 

\begin{equation}
    p_{uvl}= 
\begin{cases}
    1 ,& \text{if } class(u(i,j), v(i,j)) = l\\
    0,              & \text{otherwise}
\end{cases}
\end{equation}
In our case, we are using a panorama image for mapping, which means $p_{uvl}$ is obtained for each small rotation of the agent $\theta$ that is taken to complete a full rotation $2\pi$. Therefore $p_{uvl}$ at a relative rotation of the agent becomes $p_{uvl}^{\theta}$. Let $R(\theta)$ be the transformation matrix that rotates a competing element by the angle $\theta$. Then the final $(1\times n)$ dimensional grid element of the projected map with grid index $u(i,j), v(i,j)$ is given by:
\begin{equation}\label{node_elem}
    p_{uvl}= \frac{\theta}{2\pi}\sum_{\theta} R(\theta)p_{uvl}^{\theta}
\end{equation}

\subsection{Graph convolution for approximate projection map refinement}
Our proposed graph convolution framework is applied to approximate projection map to obtain the near perfect representation of the environment around the agent. Each grid node around the agent is classified as either a navigable space, obstacle, target object, or unmappable (beyond wall). Figures~\ref{map1},~\ref{map2},~\ref{map3},~\ref{map4} presented demonstrate collected input vision data by the agent, approximated projection maps and the corrected map after application of graph convolution algorithm. A quantitative analysis of how much better of a better map can be obtained than pure approximate projection is also presented in the table~\ref{mapping-table}.

\subsection{Extracting semantic object priors}
\begin{figure}
  \centering
  \includegraphics[width=0.50\textwidth]{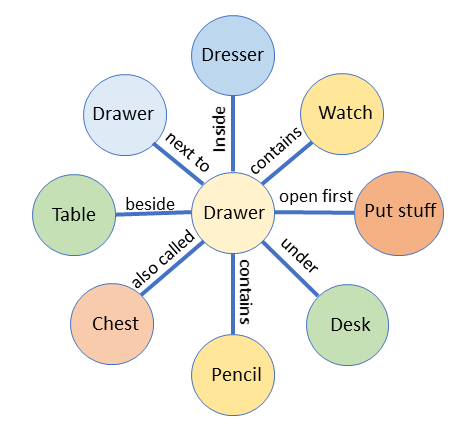}
  \caption{\textit{Illustrative example for knowledge graph for a Drawer object extracted from language corpus of instructions involving drawer.}}
  \label{fig:knowledge_graph}
\end{figure}

The Language understanding module also provides a method to construct semantic relationships in between common household objects with regards to fulfillment of common household tasks. This kind of ``common sense" relationships can be obtained in form of a densely connected knowledge graph a small part of which is illustrated as an example in figure~\ref{fig:knowledge_graph}. The example shows semantic relationships of the household object ``Drawer" with several other household objects all of which are specified as nodes of the graph and edges connecting them describe the relationship between them. We propose a simple technique for automated extraction of these relationship patterns purely from language instruction data. Let $\mathcal{L}=\{l_1,l_2,l_3,\dots,l_n\}$ denote a corpus of language instruction data involving execution of tasks in indoor household environments. Let each $l_i = \{s_{i1},s_{i2},\dots,s_{i12}\}$ contain a list of slot labels ($s_{ij}$) for each word in the instruction (see figure~\ref{fig:slot_filling} for list of 12 slot labels). Our algorithm makes two passes over the corpus $\mathcal{L}$. In the first pass, it identifies unique objects mentioned in the language by querying the slot label for ``target-obj" and ``refinement-obj" which are extracted by BERT. Each unique object is added as a node in the graph. For the same instruction, it also queries the slot labels ``action-n-navi" and ``action-desc" to determine the behavior of the agent expected towards the object such as whether it is generally something to be picked up (eg-pen), or something which serves as a big landmark for the agent to navigate (eg-bed). The words obtained by querying the slot labels for the instruction are added to a list of possible edge relationships which would be added later to connect the nodes of the graph. Same query can be made for the slot labels ``refinement-rel" and ``target-rel" to determine the location relationship (eg-beside,above,below,together) of the ``target-obj" and the ``refinement-obj". The words obtained by querying the slot labels are again added into the list of possible edge relationships. In this way, after the first pass, the list of all nodes of the graph is obtained along with the possible edge relationships. However, these edge relationships can be highly repetitive, for eg- the words ``pick",``grab",and ``take" describe the same relationship in the sentence-``pick up the cellphone from the desk" for the node pair- ``cellphone" and ``desk". This is why, we need to condense the edge relationships into as few distinct patterns as possible. For this, we use combination of several word embeddings and compare pairwise similarity. Although very rudimentary, this technique tends to work best for our scenario. A threshold value is determined based on repeated trials for a fixed corpus $\mathcal{L}$ and relationship pairs that have similarity values above threshold are renamed into a common name (eg-``pick"=``grab"=``take"). In the second pass of the algorithm, the nodes are then connected to each other through the refined relationships. 

\begin{figure*}
    \centering
    \includegraphics[width=1.0\linewidth]{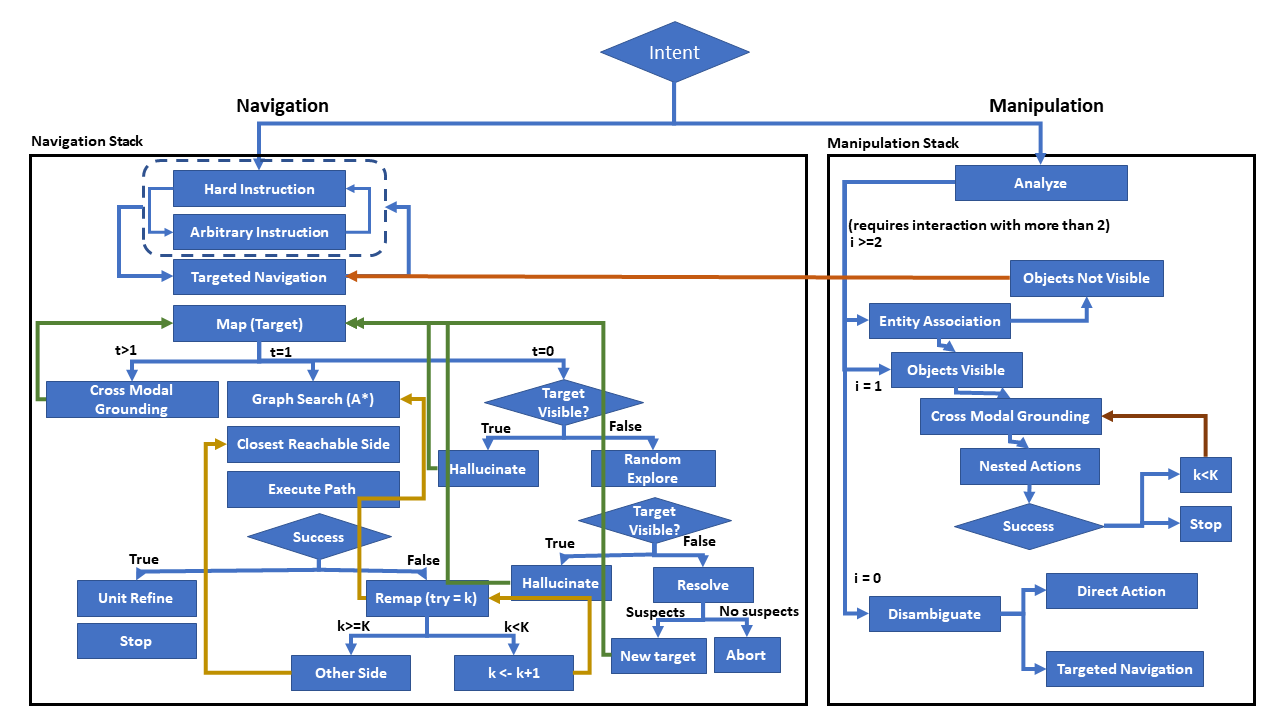}
    \caption{\textit{High level schematic of planning algorithm}}
    \label{ruletree}
\end{figure*}

\subsection{Resolving ambiguity in object descriptions and generalizing to unknown objects}
Image instance segmentation is used alongside depthmaps and RGB images in our algorithm. However ambiguous natural language descriptions of a task can occur when an agent fails to recognize a different name for the same object visible in the scene. Ambiguous description can also occur when a completely new object category (to a trained semantic segmentation model) is expected to be related to a ``target-object" mentioned in the language instruction. We propose a 2-way disambiguation for these cases. This technique relies on the agents ability to connect to an online image database (eg-google image search) and download images corresponding to a specified keyword. Lets say for example, a language instruction specifies the agent- ``Go to a dogbed", but the image semantic segmentation model recognizes the object ``beanbag", and ``beanbag" is visible to the agent. The 2-way disambiguation can be briefly summarized below:
\begin{enumerate}
    \item Disambiguate by pixel comparison
    \begin{enumerate}
        \item Download $N$ images for the query object specified by the user in natural language (eg-``dogbed")
        \item Let there be $K$ objects visible in the scene. For all objects visible in the scene, extract the group of pixels in the image corresponding to its object category as recognized the the semantic segmentation framework.
        \item Forward pass $N$ images for query object through pretrained ResNet and obtain the feature vector as an embedding for the image. Call it $N_f= \{ N_{f1},\dots N_{fN}\}$. Forward pass $K$ images for visible objects through pretrained ResNet to obtain $K_f = \{ K_{f1},\dots K_{fK}\}$
        \item Calculate pairwise cosine similarity between each element of $N_f$ and $K_f$, and chose $K_{fi}$ that has the highest value with all the elements of $N_f$. The disambiguated object is the object corresponding to $K_{fi}$.
    \end{enumerate}
    \item Disambiguate by label comparison
    \begin{enumerate}
        \item Download $N$ images for the query object specified by the user in natural language (eg-``dogbed")
        \item Let there be $K$ objects visible in the scene. For all objects visible in the scene, query the name for the object category as recognized the the semantic segmentation framework.
        \item Download 1 image for each recognized object category with the query name as the segmentation class name(eg-``beanbag"), downloading a total of $K$ images.
        \item Forward pass $N$ images for query object through pretrained ResNet and obtain the feature vector as an embedding for the image. Call it $N_f= \{ N_{f1},\dots N_{fN}\}$. Forward pass $K$ images for visible objects through pretrained ResNet to obtain $K_f = \{ K_{f1},\dots K_{fK}\}$
        \item Calculate pairwise cosine similarity between each element of $N_f$ and $K_f$, and chose $K_{fi}$ that has the highest value with all the elements of $N_f$. The disambiguated object is the object corresponding to $K_{fi}$.
    \end{enumerate}
\end{enumerate}

\subsection{Modular planning for execution of low level actions from high level behaviors}

For the final task for executing natural language instructions through low level actions, we propose a generic condition based planner. We proposed these conditions based on extensive study of 25000+ natural language instructions and incorporated human like common sense reasoning through the conditions. Most of the components of the planer leverage specialized modules that can be trained independently using state of art deep learning models or customized using additional user specified domain knowledge. Because each of the tasks of visual perception, language understanding, as well as cross modal grounding and disambiguation operate and can be trained independently, the bulk of planning boils down to choosing a high level agent behavior at a time step which is executed as a sequence of low level actions over several time steps. Here we provide a complete visualization of all the high level behaviors that are available to the agent in figure~\ref{ruletree}. Popular reinforcement learning techniques can be easily integrated to close the loop by choosing the high level action given as input state- outputs of mapping, language understanding and cross modal grounding components, and execute a sequence of low level actions, however integrating reinforcement learning based techniques is left as future work. The planner consists of two main modules- the navigation stack and the manipulation stack. Any instruction that has both navigation and manipulation components can be executed part by part as navigation and manipulation instructions. In the navigation stack, input navigation instruction can specify the agent to execute either a hard instruction such as -``turn around", ``walk 2 steps forward", arbitrary instructions, such as- ``walk forward a few steps" or targeted navigation such as- ``walk over to the desk".  These are high level behaviors which are exactly determined by querying slot labels corresponding to the parse of the instruction by the language understanding module. Hard and arbitrary instructions can be executed by direct application of low level commands as extracted by querying slot labels of the instruction. On reaching targeted navigation, the agent utilizes the mapping module which relies on graph convolution filtering to provide an explicit BEV map of agent surroundings. However the number of targets $t$ can be greater than 1, equal to 1, or even 0 when it is beyond depth perception or not visible to the agent, or is assigned a different object category (ambiguous instruction). When $t>1$ the cross modal grounding module is invoked which matches groups of pixels in image to object descriptions in language to reduce the number of targets to 1. If $t=1$, a straightforward A* planning can be invoked to enable the agent to reach the target location. If there is an obstruction not accounted for by the mapping module (or moving objects), path planning may fail to execute, in that case, a small allowance is provided to the agent for a fixed number of collisions during which it remaps its surroundings. Popular techniques like Kalman filters can also be used to fuse different map readings over time. If the number of allowances is exceeded, the agent can try to face the target in a different direction (all objects have 4 different planning targets because of box like BEV approximation). Finally, when $t=0$, the first check is done to ensure the object is visible in RGB frames (if its too far for depth perception). If visible in RGB, the agent makes a move in the general direction of the object considering the mapped navigable space around it. If the object is not RGB visible, a small neighborhood random exploration is done to make sure the object has not been cut off from visibility due to walls or other obstructions. If the object then becomes visible, mapping module is called again to receive a new map. However if the object is still not visible, the disambiguation module (``Resolve") is invoked, which relies on an internet connection to an online image search database to disambiguate objects and provide a valid target that matches language description. After navigation succeeds, a unit refinement step is executed in which the agent takes a unit step in the direction of the object which maximizes the segmented area of the object. 

For the manipulation stack, the slot labels for the instruction is analyzed to decide whether the number of interaction objects $i>=2$ (for example-``Place the book on the table to the right of the lamp") or $i=1$ (for example- ``turn on the light") or $i=0$ (for example -``gaze upwards"). In case $i>=2$ an entity association is done based on slot labels which places objects in a hierarchy of interaction (for example- table is level 0, lamp is level 1, book is level 2). After that for both $i>=2$ or $i=1$, a check is done to make sure the objects mentioned in language are visible in RGB frame. If multiple objects of same category are present, cross modal grounding module is invoked, however if none of the objects are found to match with object categories for image segmentation, then disambiguate module is invoked. Finally after making sure the objects are visible,grounded and disambiguated, low level actions such as ``pick(book,table)" or ``turn-on(lamp)" can be executed.

\section{Conclusion}
A major reason for the significant improvement shown by our modular framework compared to the seq-to-seq models can be attributed to the disentangled manner in which important features of each modality are extracted and combined to obtain high level behaviors. Many end to end frameworks, even with sophisticated attention mechanisms, struggle to learn these disentangled representations. This is primarily because of the inherent many-to-one nature of the problem - a language description and a sequence of visual observations can jointly lead to multiple long sequences of action outputs. Therefore, a modular approach that encodes human-level expert semantic knowledge and a generalized understanding of word-level semantics can pave the way to train agents with more human like understanding of the tasks and hence, better success rates.

\section{Broad Societal Impact}
 With improved quality of living and better healthcare, we can expect a major section of older population as well as the disabled requiring assisted living in the future benefiting from these human robot interaction technologies. Better techniques for improved navigation and instruction following as provided in our paper can help design the next generation of interactive robots that are better suited to interact socially and understand instructions provided to them in natural language. 

{\small
\bibliographystyle{ieee_fullname}
\bibliography{ref}
}

\end{document}